\documentclass[conference]{IEEEtran}
\IEEEoverridecommandlockouts
\usepackage{amsmath,amssymb,amsfonts}

\usepackage[linesnumbered,ruled,commentsnumbered]{algorithm2e}

\usepackage{graphicx}
\usepackage{subfigure}
\usepackage{stfloats}
\usepackage{textcomp}
\usepackage{balance}
\usepackage{xcolor}
\usepackage{enumitem}
\usepackage{caption}
\usepackage[colorlinks=true,
            linkcolor=blue,
            anchorcolor=black,
            citecolor=black,
            ]{hyperref}
\usepackage{booktabs}
\usepackage{cite}
\usepackage{url}
\usepackage[margin=1.9cm]{geometry}
\def\BibTeX{{\rm B\kern-.05em{\sc i\kern-.025em b}\kern-.08em
	T\kern-.1667em\lower.7ex\hbox{E}\kern-.125emX}}

\DeclareRobustCommand*{\IEEEauthorrefmark}[1]{%
    \raisebox{0pt}[0pt][0pt]{\textsuperscript{\footnotesize\ensuremath{#1}}}}

\makeatletter
\def\ps@IEEEtitlepagestyle{%
  \def\@oddhead{%
    \hbox{%
      \hspace*{0.3\textwidth} 
      This article has been accepted by \href{https://doi.org/10.1109/ICRA57147.2024.10610488}{ICRA2024} 
    }%
  }%
  \def\@evenhead{\@oddhead}%
  \def\@oddfoot{}%
  \def\@evenfoot{}%
}
\makeatother

\begin{document}
\newgeometry{left=1.9cm, right=1.9cm, top=2.54cm, bottom=1.9cm}
\title{History-Aware Planning for Risk-free Autonomous Navigation on Unknown Uneven Terrain
}

\author{Yinchuan Wang$^{1*}$, Nianfei Du$^{1*}$, Yongsen Qin$^{1}$, Xiang Zhang$^{1}$, Rui Song$^{1,2}$, Chaoqun Wang$^{1}$%

\thanks{This work was supported in part by the National Natural Science Foundation of China under Grant No. 62103237}
\thanks{\IEEEauthorrefmark{1}Yinchuan Wang, Nianfei Du, Yongsen Qin, Xiang Zhang and Chaoqun Wang are with the School of Control Science and Engineering, Shandong University, \#17923, Jingshi Road, Jinan, Shandong Province, China. Email: \tt \{sdwyc, nianfei.du, yongsen.qin, sucro\_zhangxiang\}@mail.sdu.edu.cn, \{chaoqunwang, rsong\}@sdu.edu.cn}%
\thanks{\IEEEauthorrefmark{2}Rui Song is with Shandong Research Institute of Industrial Technology \#17923, Jingshi Road, Jinan, Shandong Province, China.}
\thanks{$^{*}$The first two authors contributed equally to this work.}
}

\maketitle
\begin{abstract}
It is challenging for the mobile robot to achieve autonomous and mapless navigation in the unknown environment with uneven terrain. In this study, we present a layered and systematic pipeline. At the local level, we maintain a tree structure that is dynamically extended with the navigation. This structure unifies the planning with the terrain identification. Besides, it contributes to explicitly identifying the hazardous areas on uneven terrain. In particular, certain nodes of the tree are consistently kept to form a sparse graph at the global level, which records the history of the exploration. A series of subgoals that can be obtained in the tree and the graph are utilized for leading the navigation. To determine a subgoal, we develop an evaluation method whose input elements can be efficiently obtained on the layered structure. We conduct both simulation and real-world experiments to evaluate the developed method and its key modules. The experimental results demonstrate the effectiveness and efficiency of our method. The robot can travel through the unknown uneven region safely and reach the target rapidly without a preconstructed map. 
\end{abstract}

\section{Introduction}
\label{introduction}

The rapid development of mobile robots has unlocked their potential in a wide range of applications\cite{app1,app2,app3}. More and more robot platforms exhibit autonomous navigation capabilities in structured environments relying on preconstructed maps. 
It is still challenging to achieve reliable navigation in unstructured environments without a prior map, which is known as mapless navigation \cite{mapless_nav}. 


Recent years have witnessed the emergence of different efforts towards mapless navigation\cite{mapless1, mapless2}, which devote themselves to empowering robots to autonomously navigate in unknown environments leveraging limited perceptual capabilities. Despite the rapid progress, autonomous mapless navigation on uneven terrain is still left unsolved. With regard to this topic, one may encounter more challenges. Oftentimes there is no obvious gap between the undulated ground and obstacles in uneven terrain environments, making the traversability analysis a difficult task. Consequently, it is hard to explicitly obtain the hazardous area in the environment to achieve safe navigation. Moreover, quite a few outdoor uneven terrain environments either lack representative features or suffer from feature degradation problems. Without enough supportive clues, it is hard for the decision-making of where to go to efficiently reach the destination across the unknown. 

\begin{figure}[tp]
	\centering
	\setlength{\abovecaptionskip}{-5pt}
	\setlength{\belowcaptionskip}{-10pt}
	\includegraphics[scale=0.28]{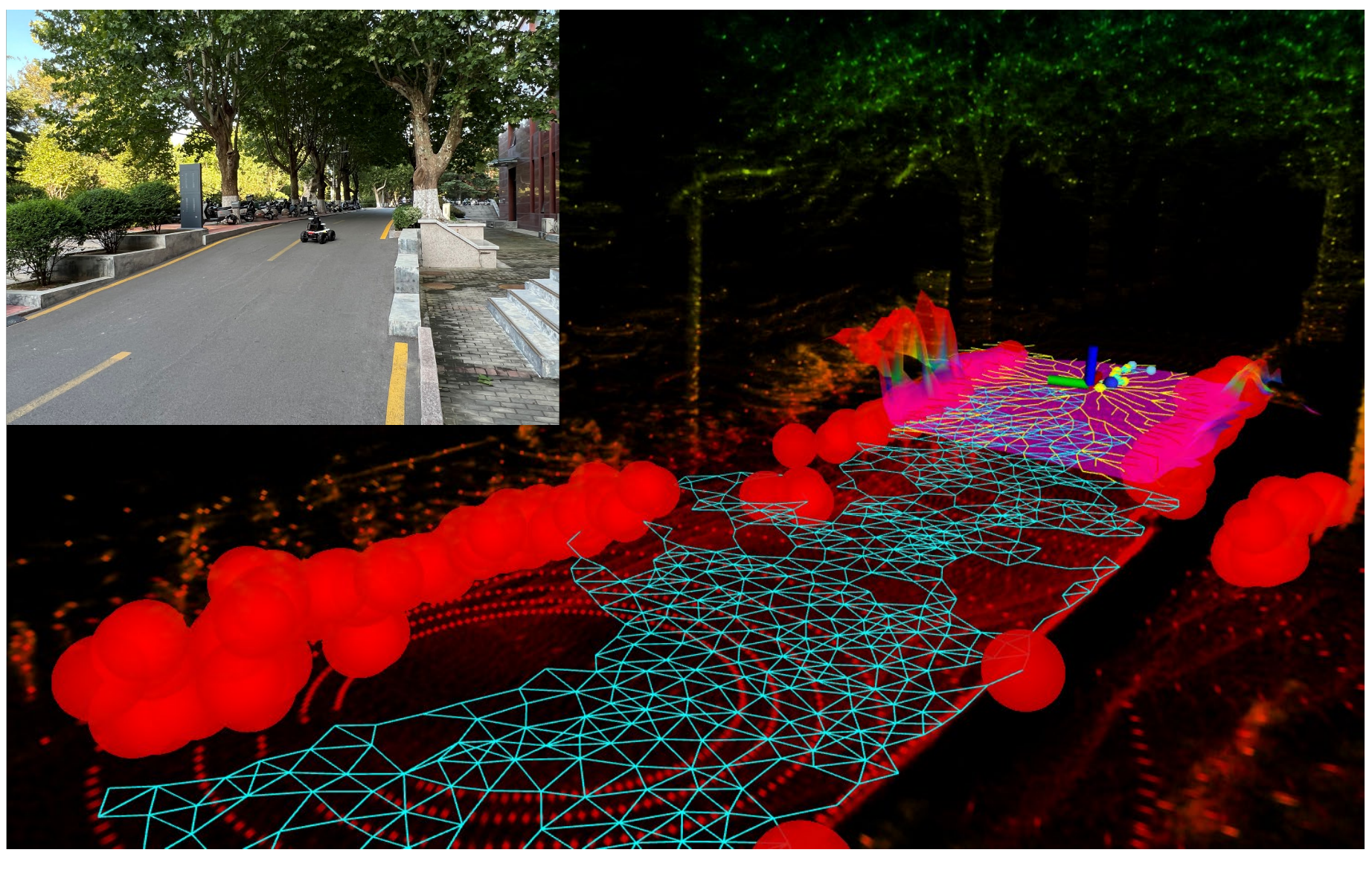}
        \vspace{0.2cm}
	\caption{The application of the developed pipeline in real-world environments. A local tree and a global graph are maintained along with the navigation.}
        \vspace{-0.25cm}
	\label{fig: topFig}
\end{figure}


In this paper, we present a unified pipeline for risk-free and efficient mapless navigation in unknown uneven environments. This method works at two different levels. At the local level, we maintain a Hazard-aware Dynamic RRT (HD-RRT), which can offer not only the navigation subgoal, but also the route to the target. Besides, this tree structure helps mark the hazardous area in the uneven terrain environment. 
This tree structure is dynamically changed and pruned to maintain its lightweight. To keep the memory of the explored environment, we maintain a graph structure at the global level, which can also provide the leading subgoal and the navigation route. To determine where to go during the mapless navigation, we design layered heuristic cost functions, in which the ingredients are efficiently provided by the tree and graph structures. In summary, the contributions of the developed pipeline are as follows
\begin{itemize} 
  \item We develop a lightweight HD-RRT, which can efficiently lead the environment exploration and mark explicitly the hazardous regions in the environment. 
  \item We maintain a graph structure to represent the explored environment to keep the navigation history, which alleviates the need for large-scale maps. 
  \item A decision-making module is designed based on the data rapidly provided by the HD-RRT and the graph. It can effectively lead the mapless navigation. 
\end{itemize}
Both the simulation and real-world experiments are conducted and results support the claims of presented framework. 
\newgeometry{left=1.9cm, right=1.9cm, top=1.9cm, bottom=2.0cm} 
\noindent The video demonstration of conducted experiments is available at link\footnote{Video demonstration: \href{https://www.youtube.com/watch?v=znzEKKJr06w}{www.youtube.com/watch?v=znzEKKJr06w} }, while the code is open-sourced in the repository\footnote{Code: \href{https://github.com/sdwyc/history-aware-planner}{https://github.com/sdwyc/history-aware-planner} }

The remainder of this article is structured as follows: 
Sec. \ref{related_work} discusses the related research. Sec. \ref{preliminary} presents the preliminaries and describes the problems, followed by the system pipeline in Sec. \ref{system overview}. Subsequently, the proposed method is introduced in  Sec. \ref{methodology}. The conducted experimental evaluation is presented in Sec. \ref{experiment}. We draw conclusions and discuss the future work in Sec \ref{conclusion}.

\section{Related Work}
\label{related_work}

Mapless navigation has long been a hot research topic over the past few decades\cite{drl_uneven1, attention2, new_reward}. The absence of prior maps raises great uncertainty to the planning and decision making in the navigation. For years, a considerable portion of research has focused on path planning in partially or totally unknown environments \cite{ga_planning, swarm1, faster}. They generally leverage optimization methods to minimize their object function and output an accessible path to the target, while maintaining an incrementally constructed map. For the decision-making part, \cite{map_predict, jian2022putn} analyze the geometrical features to sense the environments and make optimal decisions based on those features. However, these methods are inefficient since they often maintain an incremental established global map and need large computation overhead. 


With the soaring in popularity of deep reinforcement learning (DRL) techniques, more researchers have noted their potential for solving path planning and decision-making in mapless navigation tasks. General DRL-based methods \cite{zhu2017target, DRL2, DRL3, DRL4} exploit limited sensing information to train the model and finally output an optimal policy in the environments without prior maps. 
In addition, there have been efforts \cite{imitation1, imitation2} that combine DRL and imitation learning (IL) methods to imitate human expert behaviors to make complex decisions for mapless navigation. These approaches could significantly increase the generalization ability in various environments. In large-scale or unstructured environments, the performance of these methods may be drastically deteriorated due to the increase of the decision-maksing space. 

Regarding the realm of robot navigation without prior maps, structured and uneven environments exhibit significant disparities. Indeed, there have been attempts that explore DRL and their variants to solve navigation problems in uneven terrain \cite{attention1, drl_uneven2, drl_uneven3}. It must be noted that these methods may not be applicable when the environment is too large and not be available when the terrain feature degenerates. To achieve reliable navigation performance, various frameworks \cite{jian2022putn,warnke2020towards,dergachev20222} have been developed in recent years with separate focus on either traversability analysis, path planning, or motion safety.  The developed framework provides a unified view towards the mapless navigation in uneven terrain. The traversability and safety analysis is directly incorporated to the decision-making and path-planning framework, which offers a more compact solution to lead the mapless navigation.

\section{Preliminaries}
\label{preliminary}
The course of autonomous mapless navigation can be described as an incremental exploration process where a series of subgoals $\mathcal{S}$ are determined and used for leading the navigation. The problem is hereby formulated by
\begin{equation}
\label{mapless_description}
\pi^{*} \doteq \underset{\{S_i \in \mathcal{S}\}}{\textit{argmin}} \sum  F(\mathbf{x}_{r} \rightarrow
 S_i),
\end{equation}
where $ F(\cdot)$ is the utility function evaluating the cost of moving the robot from its current location $\mathbf{x}_{r}$ to a subgoal $S_i$. 
$\pi^{*}$ represents selection policy of those subgoals.
Eq. \ref{mapless_description} aims to find an optimal strategy $\pi^{*}$ that generates a combination of a series of subgoals $\mathcal{S}$ before finally reaching the target $S_{+}$.


We employ the grid map\cite{elevation_mapping} $\mathcal{M}_{L}$ to describe the uneven terrain around the robot. $\mathcal{M}_{L}$ is a $W \times H$ fix-sized and consists of a set of grids $m_i$. 
We maintain the tree structure $\mathcal{T}=\{\mathcal{V}_{\mathcal{T}}, \mathcal{E}_{\mathcal{T}}\}$ and a graph structure $\mathcal{G}=\{\mathcal{V}_{\mathcal{G}}, \mathcal{E}_{\mathcal{G}}\}$, where $\mathcal{V}$ and $\mathcal{E}$ represent the nodes and edges, respectively. 


\begin{figure}[tp]
	\centering
	\setlength{\abovecaptionskip}{-5pt}
	\setlength{\belowcaptionskip}{-10pt}
	\includegraphics[scale=0.38]{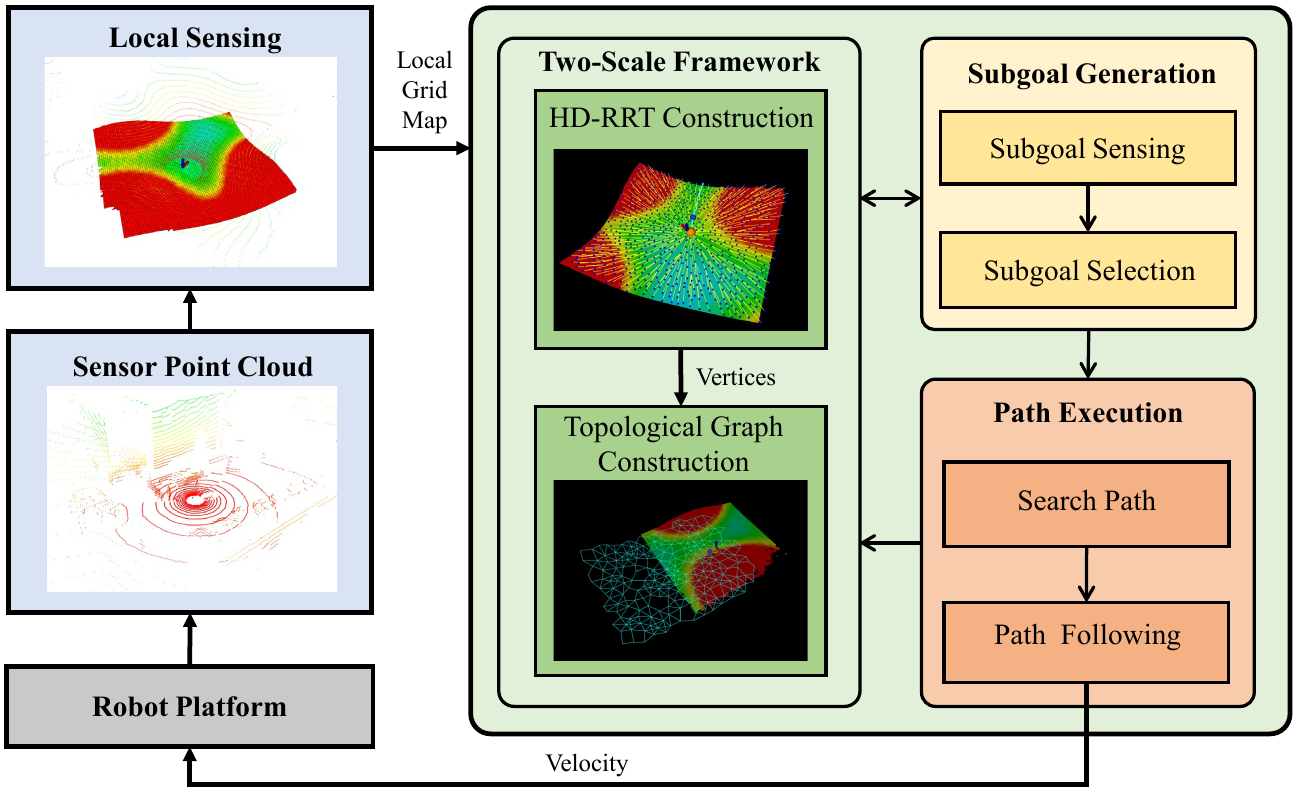}
        \vspace{0.2cm}
	\caption{System diagram of the developed mapless navigation framework.}
        \vspace{-0.3cm}
	\label{fig: system}
\end{figure}
\section{System Overview}
\label{system overview}

The developed pipeline is shown in Fig. \ref{fig: system}. 
The input of the system is LiDAR point cloud generated by a ground robot platform. 
The point cloud is used for the local sensing process and outputs a 2.5D local grid map $\mathcal{M}_L$. The tree structure is built in real-time on the local grid map, on which the hazardous area is marked by the tree extension. We utilize the nodes on the tree to construct a global graph, which maintains the exploration history. Both the tree and the graph structure can provide the necessary subgoals for leading the navigation. They are evaluated and a subgoal will be determined. After that, one can efficiently query a path to the subgoal either from the tree or the graph. Then the path is optimized with an off-the-shelf trajectory optimization method (e.g.,\cite{local_planning} ) for the robot to execute. This course repeats until the robot reaches the final destination. 


\section{Methodology}
\label{methodology}
\subsection{Hazard-aware Dynamic RRT (HD-RRT) }

Different from the 2D or 3D cases where the obstacle and collision-free area are explicitly distinguishable, there is no well-defined obstacle for robot navigation on rough terrain. The feasible region for the 2.5D navigation measures whether it is traversable in light of the vehicle's climbing ability and tip-over stability. For grid map based representation, it is straightforward to evaluate each grid $m_i$ and its close neighbor $m_i^{nbr}$ who shares one grid boundary, i.e.,
\begin{equation}
    \mathcal{F}(m_i \leftrightarrow \forall m_i^{nbr}), \forall m_i \in \mathcal{M}_{L}, m_i \cap m_i^{nbr} \neq \emptyset.
\label{eq: 2}
\end{equation}
Here $\mathcal{F}(\cdot)$ is the traversable evaluation function, which incurs either higher computational cost or lower accuracy with different map resolutions. 
To this end, we utilize the sampling-based approach and develop a dynamic RRT method to release the constraint in Eq. \ref{eq: 2}. The extension of the developed tree structure is depicted in Alg. \ref{alg: DT-RRT expansion}. The $\mathbf{SampleNewNode}(\cdot)$ function is to sample uniformly within the $\mathcal{M}_{L}$ and to find a new candidate node, e.g., $m_j^{nbr*}$, where $m_j^{nbr*}$ is not necessarily an ambient grid around $m_j$. One just needs to check $\mathcal{F}(m_i \leftrightarrow m_j^{nbr*})$ instead of all the ambient neighbors. Moreover, connecting a distant node from the current location may contribute to the exploration of the state space and therefore facilitate the path planning. 

\begin{algorithm}[tp]
\caption{HD-RRT Expansion and Pruning}
\label{alg: DT-RRT expansion}
\LinesNumbered
\KwIn{target point $S_{+}$, robot position $\mathbf{x}_{r}$, $\mathcal{M}_L$}

$\mathcal{V}_{\mathcal{T}} \gets \mathbf{x}_{r}$; $\mathcal{E}_{\mathcal{T}} \gets \emptyset$; 
$\mathcal{T} = (\mathcal{V}_{\mathcal{T}}, \mathcal{E}_{\mathcal{T}})$ 


\While{$S_{+} \ not \  found$}{
    $m_j^{nbr*} \gets \mathbf{SampleNewNode}(\mathcal{M}_L, \mathcal{T})$ \\
    \eIf{$\mathbf{CheckFeasibility}(\mathcal{E}_{new})$}{
        $\mathcal{V}_{\mathcal{T}} \gets \mathcal{V}_{\mathcal{T}} \cup \{m_j^{nbr*}\}$; $\mathcal{E}_{\mathcal{T}} \gets \mathcal{E}_{\mathcal{T}} \cup \{\mathcal{E}_{new}\}$
    }
    {
        $\mathbf{UpdateSaturatedNode}(m_j)$
        $\mathbf{MarkHazard}(m_j \ is \ saturated)$
    }
    \If{$\mathbf{IsRobotMove}(\mathbf{x}_{r})$}{
        $\mathbf{PruneTree}(W \times H, \mathcal{V}_{\mathcal{T}}, \mathcal{E}_{\mathcal{T}}) $ \\
        $\mathbf{ChangRoot}(\mathbf{x}_{r}) $
    }
}
\end{algorithm}

However, since the vehicle can not be simply regarded as a mass point, it is still challenging to get $\mathcal{F}(m_i \leftrightarrow m_j^{nbr*})$. 
For one thing, one has to carefully check the slope of the segment $m_i \leftrightarrow m_j^{nbr*}$ with a suitable resolution to match the nearby terrain slope as closely as possible.
For another thing, one must consider the tip-over stability when the vehicle follows the path segment, which is also a computationally expensive simulation task\cite{traversability_mapping}. To mitigate this issue, we develop a feasible checking method. The segment $m_i \leftrightarrow m_j^{nbr*}$ is divided into $n$ smaller meta segments $\epsilon_{i}, i=1,2,\dots,n$. For each meta segment $\epsilon_{i}$, its gradability $\ell$ is calculated by
\begin{equation}
    \ell(\epsilon_{i}) = \frac{\triangle h_{i}}{\triangle l_{i}}, 
\end{equation}
where $\triangle h_{i}$ and $\triangle l_{i}$ represent the ground elevation differences and length of $i$-th meta edge $\epsilon_{i}$ respectively. The feasibility of the segment is then determined by
\begin{equation}
\label{eq: checkFeasibility}
\begin{aligned}
\mathbf{\mathcal{F}}(m_i \leftrightarrow m_j^{nbr*}) =
\left\{\begin{matrix}
 False  & \text{if} \left\{\begin{matrix}
\exists \ell(\epsilon_{i}) > \alpha \ \text{or}  \\
\sum \ell(\epsilon_{i})>\beta
\end{matrix}\right.   \\
 True   &\text{otherwise}.
\end{matrix}\right.
\end{aligned}
\end{equation}
Here $\alpha$ and $\beta$ are two thresholds where $\alpha$ represents the limitation of the vehicle's gradeability and $\beta$ is set as a reference for the overall flatness.  
The former threshold is utilized to reject the path that is too steep for the robot to follow while the latter threshold is expected to ensure the smoothness of the generated path. This checking is indicated at Line 4 in Alg. \ref{alg: DT-RRT expansion}.

\begin{figure}[tp]
	\centering
	\setlength{\abovecaptionskip}{-5pt}
	\setlength{\belowcaptionskip}{-10pt}
	\includegraphics[scale=0.46]{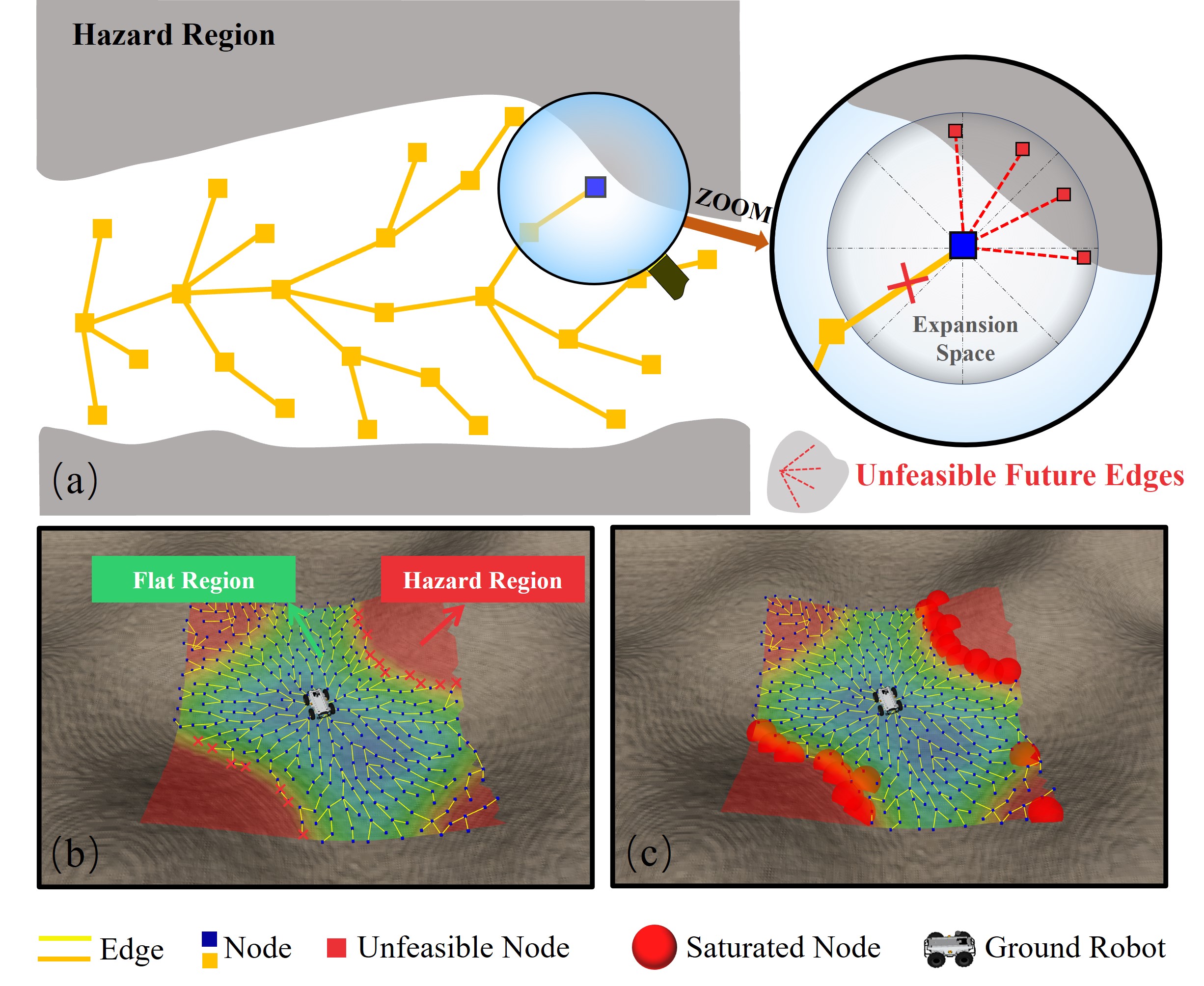}
        \vspace{0.2cm}
	\caption{Illustration of the hazardous region generation. In (a), the blue node is checked. It may incur unfeasible edges represented by red virtual lines within an eight-sector space because it is close to hazardous regions. If it is saturated then it will be marked on the map, as shown in (b). (c) showcases the hazardous regions estabilish based on the saturated nodes to avoid unsafe navigation maneuvers.}
        \vspace{-0.2cm}
	\label{fig: saturated_node}
\end{figure}

The feasibility-checking method in Eq. \ref{eq: checkFeasibility} avoids the intractable simulation of tip-over possibility. To guarantee navigation safety, we exploit the tree structure to incrementally and explicitly mark the unsafe region on uneven terrain. We introduce the saturated node to avoid $\mathcal{T}$ expanding towards hazardous regions, like walls and precipitous slopes. We associate each node on the tree with a saturated vector $\mathbf{V}_{m_j}$. Initially, $\mathbf{V}_{m_j}^{\top}=[\mathbf{0}_{1 \times 8}]$.
For a node $m_j$ that fails the feasibility checking in Eq. \ref{eq: checkFeasibility}, the corresponding $k$-th element in $\mathbf{V}_{m_j}$ will be number 1. Here $k$ is calculated by $k= Int[\angle \overrightarrow {m_jm_j^{nbr*}}/(2\pi/8)]$, where the $Int[\cdot]$ represents the rounding operation. The $\mathbf{V}_{m_j}$ will be updated when a sample node $m_j^{nbr*}$ fails to connect $m_j$ through Eq. \ref{eq: checkFeasibility}. The node $m_j$ is regarded as a saturated node when
\begin{equation}
\label{eq: saturate vector}
\begin{aligned} 
&\left \| \mathbf{v}_{n}  \right \|  \ge N_{s}, N_{s} \in \{1,2,...,8\},\\
\end{aligned}
\end{equation}
where $N_{s}$ is the saturated metric threshold. Fig. \ref{fig: saturated_node} illustrates the evaluation process of a saturated node. The node $m_j$ actually checks its 8 neighbor nodes through sampling towards tree building. If $m_j$ is a saturated node, it will be pruned by the tree $\mathcal{T}$ together with the associated edge. 


It is noteworthy to realize that the saturated node actually indicates the hazardous area on the uneven terrain, as the red cross marker illustrated in Fig. \ref{fig: saturated_node}(b). To explicitly mark the hazardous regions, we inflate the saturated node by the extension radius \cite{inflation} of RRT*, as showcased by the red bubbles in Fig. \ref{fig: saturated_node}(c). The regions covered by these bubbles are marked as hazardous areas. Intuitively, the saturated nodes could prevent $\mathcal{T}$ from expanding inside hazardous regions, which guarantees the feasibility of tree expansion space. As the robot moves, the tree is extended continuously and the hazardous regions can be marked incrementally on the map. 

\begin{figure}[tp]
	\centering
	\setlength{\abovecaptionskip}{-5pt}
	\setlength{\belowcaptionskip}{-10pt}
	\includegraphics[scale=0.4]{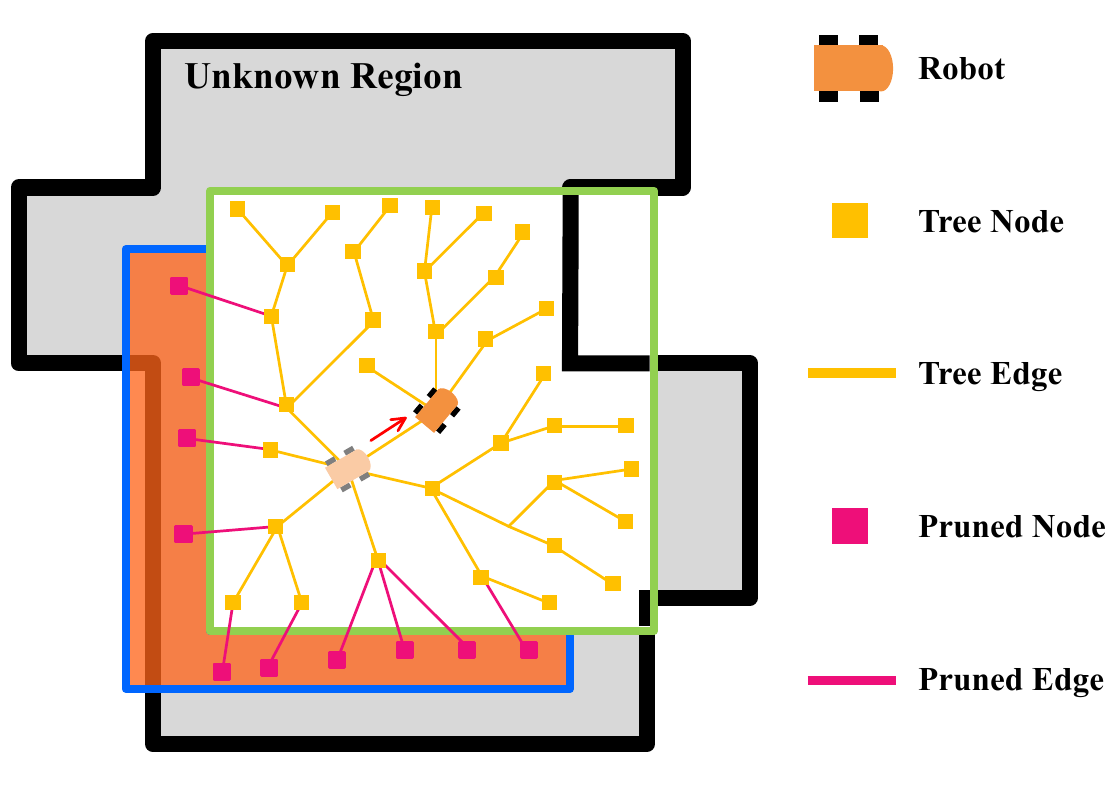}
        \vspace{0.4cm}
	\caption{The process of pruning nodes and edges. The green line is the current local boundary while the blue line is the boundary of $\mathcal{M}_{L}$ at the last-moment.}
        \vspace{-0.25cm}
	\label{fig: dt_rrt prune}
\end{figure}

For the tree extension, since preserving the full edges consumes lots of rewiring computational resources, we develop to narrow down the tree extension in a sliding window. The root of the tree is aligned with the robot's location and changes as the robot moves. It avoids reconstructing the tree and helps efficiently explore the environment during the navigation. Besides, we prune the nodes and edges outside the moving window. As shown in Fig. \ref{fig: dt_rrt prune}, when the robot moves, the edges and nodes located at the semi-orange region will be pruned and are no longer used for planning.

\subsection{History-aware Global Topological Graph}
\label{global graph}


Although the HD-RRT can offer terrain and hazard awareness, the previous sensing memory is no longer preserved 
even valuable information might exist at previous locations.
To this end, we establish a history-aware topological graph $\mathcal{G}$  at the global level. The construction process of $\mathcal{G}$ follows closely the pipeline in probabilistic roadmap method (PRM)\cite{prm}. 
In particular, the $\mathcal{G}$ acquires new vertices $\mathcal{V}_{new}$ by HD-RRT sharing rather than by sampling state space. 
The valuable nodes $\mathcal{V}_{p}$ in HD-RRT for constructing the graph $\mathcal{G}$ mainly incorporates the following types of node

\begin{figure}[t]
	\centering
	\setlength{\abovecaptionskip}{-5pt}
	\setlength{\belowcaptionskip}{-10pt}
	\includegraphics[scale=0.28]{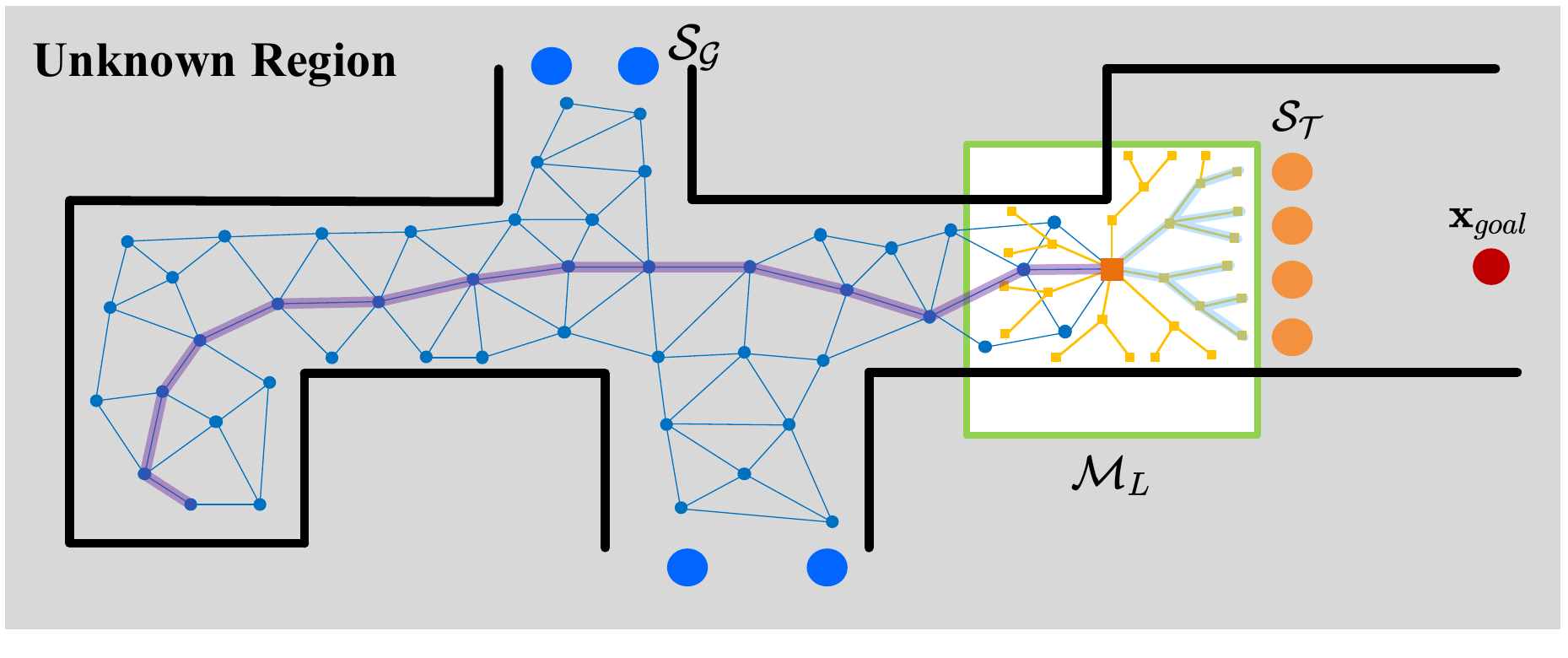}
        \vspace{-0.2cm}	
 \caption{Illustration of graph construction. The green boundary represents the range of $\mathcal{M}_{L}$. The blue lines and small circles are the vertices and edges of $\mathcal{G}$. The yellow lines and rectangles are the nodes and edges of $\mathcal{T}$. The purple line represents the trajectory of the robot and the blue shaded lines are the branches that are to be added to the $\mathcal{G}$.}
        \vspace{-0.2cm}
	\label{fig: global_graph}
\end{figure}

\vspace{-0.03cm}
\begin{itemize}
    \item \textbf{Candidate Target Nodes} $\mathcal{V}_{F}$, the leaf nodes on the tree that records the boundary information of the local window, which is specifically detailed in Sec. \ref{subgoal selection}. 
    \item \textbf{Root Nodes} $\mathcal{V}_{R}$, the previous root nodes of the HD-RRT, which indicates the safe regions and records the robot navigation history.
    \item \textbf{Waypoint Nodes} $\mathcal{V}_{W}$, the nodes on the branch of HD-RRT that connects the robot current location $\mathbf{x}_r$ and $\mathcal{V}_{F}$.
\end{itemize}
The extension of the graph $\mathcal{G}$ is shown in Fig. \ref{fig: global_graph}. It is extended along with the HD-RRT construction. 
Once a node $\mathcal{V}_{F}$ near the boundary is found, we will search for a route from the robot location to that node and obtain a branch. Then the nodes on this branch are immediately utilized for constructing the graph. The nodes will try to connect to each other with feasible branches that can be evaluated with Eq. \ref{eq: checkFeasibility}. Note that the structure of $\mathcal{G}$ remains sustainable until the whole navigating task ends. This history-aware feature could maintain the sensing memory and increase the efficiency of possible path searching.

\subsection{Subgoal Generation and Evaluation}
\label{subgoal selection}

To lead the navigation, robot consistently pursues a series of subgoals before reaching final destination. This method maintains two types of subgoals: \textbf{local subgoal} $\mathcal{S}_{\mathcal{T}}$ and \textbf{global subgoal} $\mathcal{S}_{\mathcal{G}}$. The $\mathcal{S}_{\mathcal{T}}$ is generated within the local window on HD-RRT. At a navigation instant, e.g., $t_1$, given the local tree $\mathcal{T}_{t_1}$ and denote the leaf nodes without any children node as $\mathcal{V}_l^{t_1}$. For a node $m_k^{t_1} \in \mathcal{V}_l^{t_1}$, we check the explored grids lie in the circle area around $m_k^{t_1}$ spanned tree extension radius. The explored grids are the ones in the local window offered by the grid map $\mathcal{M}_l$. We propose to determine whether the node is valuable simply by $\nabla=S_o/S_a$, where $S_o$ is the occupied area of the grid map in the circle region and $S_a$ is the total area of the circle region. For a node $m_k^{t_1} \in \mathcal{V}_l^{t_1}$, it will be a $\textbf{Candidate Target Node}$ if $\nabla \leq \delta$ at time $t_1$. The \textbf{Candidate Target Node} will be a candidate subgoal if it is always a leaf node and 
\begin{equation}
    \nabla\le \delta , t=t_1,...,t_n,
    \label{eq:subgoal}
\end{equation}
at the evaluation time $t_n$ within $\mathcal{M}_L$. Note that the $\nabla$ will only be calculated if the node is within the local window and we record the highest value $\nabla_{max}$. The candidate subgoal belongs to the set $\mathcal{S}_{\mathcal{T}}$ if its location is within the local window, otherwise it will be stored in the set $\mathcal{S}_{\mathcal{G}}$.
Given a subgoal set $\mathcal{S}={\mathcal{S}_{\mathcal{G}} \cup  \mathcal{S}_{\mathcal{T}}}$, we determine a subgoal for leading the navigation by
\begin{equation}
    S^* = \underset{S_i \in \mathcal{S}}{\textit{argmin}} \  \mathbf{Cost}(S_i) ,
\end{equation}
where the $\mathbf{Cost}(\cdot)$ is the cost value of one subgoal. In this study, we present two different cost functions to evaluate the subgoals $\mathcal{S}_{\mathcal{T}}$ and $\mathcal{S}_{\mathcal{G}}$.

\begin{figure*}[t]
\centering
\includegraphics[width=1.0\textwidth,height=6.2cm]{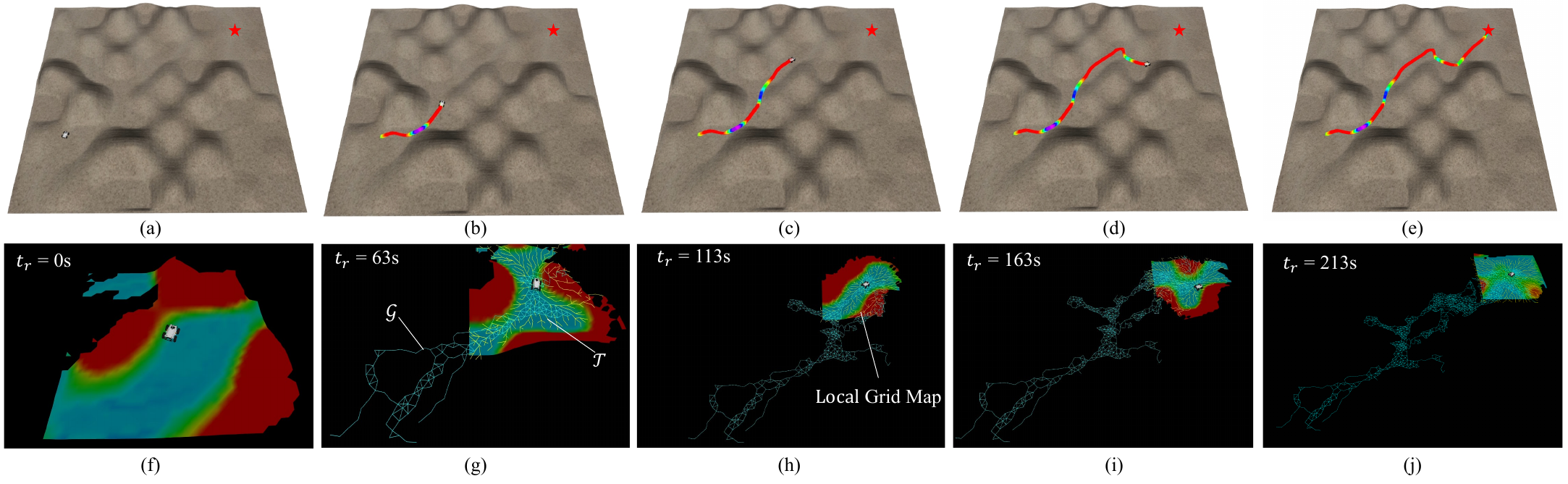}
\vspace{-0.6cm}
\caption{Total performance of the developed HAP framework. (a)-(e) show the snapshots of the navigation. The trajectory colors represent the elevation profile. (f)-(j) show the expansion of $\mathcal{T}$ and $\mathcal{G}$ during the navigation. The red star represents the target.}
\vspace{-0.5cm}
\label{fig: whole_process}
\end{figure*}  

\begin{algorithm}[t]
\caption{Decision Making}
\label{alg: decision making}
\LinesNumbered
\KwIn{$\mathcal{T}$, $\mathbf{x}_{r}$; \ \textbf{Output}: Best subgoal $S_i$}

$\mathcal{S}_{\mathcal{T}} \gets \mathbf{SubgoalGeneration}(\mathcal{T}) \  $\%$ Initialization$ \;
    \eIf{$\mathcal{S}_{\mathcal{T}} \neq \emptyset$ \  \textbf{and} \ $ N_{\bigtriangledown} \ge n_{\delta} $}
        {
            $S_i \gets \mathbf{SelectSubgoal}(\mathcal{S}_{\mathcal{T}})$\;
        }
        {
        $S_i \gets \mathbf{SelectSubgoal}(\mathcal{S}_{\mathcal{G}})$\;
        }
        \If{$\mathbf{RootChange}(\mathbf{x}_{r})$}{
        $\mathbf{Update(\mathcal{T}, \mathcal{G})}$\;\
        $\{\mathcal{S}_{\mathcal{T}}, \mathcal{S}_{\mathcal{G}}\} \gets \mathbf{SubgoalGeneration}(\mathcal{T}, \mathcal{G})$\;\
        $\mathbf{SearchAndExecutePath}(\mathcal{T}, \mathcal{G},\mathcal{M}_L)$\;
    }
\end{algorithm}

To achieve better trade-off between efficiency and safety, the cost of each $S_i \in \mathcal{S}_{\mathcal{T}}$ is computed as
\begin{equation}
\label{eq: cost function}
    \mathbf{Cost}^{\mathcal{T}}(S_{i}) = (\alpha \frac{\mathcal{T}_{{i}}}{\sum \mathcal{T}_{{i}}}+\beta \frac{\Gamma_{{i}}}{\sum \Gamma_{{i}}})e^{-\lambda \Upsilon  }+Dist(S_i, S_+),
\end{equation}
where the $\alpha$, $\beta$, $\lambda$ are the configurable weight parameters.
The $\mathcal{T}_i$ and $\Gamma_i$ are the sum of the edge length and gradient from the $\mathbf{x}_{r}$ to the subgoal goal $S_i$, respectively.  $\sum \mathcal{T}_{{i}}$ and $\sum \Gamma_{{i}}$ represent the sum of  $\mathcal{T}_{{i}}$ and $\Gamma_{{i}}$ of all local subgoals respectively. $\Upsilon$ is the sum of the angle among two consequent edges from robot location $\mathbf{x}_r$ to the subgoal $S_i$.
$Dist(\cdot)$ is a heuristic item that is denoted as the planar Euclidean distance.  The subgoal selected by Eq. (\ref{eq: cost function}) may lead to more flat path and more efficient goal searching. Only the $\mathcal{S}_{\mathcal{T}}$ is considered if there are subgoals within the local window at a decision-making instant. Otherwise, the subgoals at $\mathcal{S}_{\mathcal{G}}$ will be considered to avoid the robot trapped into a certain region. Differently, these subgoals are evaluated by 
\begin{equation}
\label{eq: global cost function}
    \mathbf{Cost}^{\mathcal{G}}(S_j) = Dist(S_j, S_+) e^{\nabla_{max}(S_j)}.
\end{equation}
This setting is utilized to select the most informative subgoal while considering its distance to the final goal.


The decision-making process is described in Alg. \ref{alg: decision making}. This process is executed repeatedly until reaching the target, with its execution dependent on the construction of $\mathcal{T}$ and $\mathcal{G}$. We just select the subgoal within the $\mathcal{T}$, evaluated by Eq. \ref{eq: cost function}, if it is not empty and it has enough candidate target nodes (see Lines 1-4), where $N_{\bigtriangledown}$ and $n_{\delta}$ are the number of candidate target node and quantity threshold. Otherwise, the subgoal will be selected within $\mathcal{G}$ and evaluated using the Eq. \ref{eq: global cost function}. Once a subgoal is determined, a path can be found on the tree or the graph and it will be immediately optimized for the robot to follow, as indicated in Line 10, Alg. \ref{alg: decision making}.

\section{Experiments and Results}
\label{experiment}

To evaluate the efficacy and efficiency of the developed framework, we conduct a series of both simulation and real-world experiments. 
\vspace{-0.3cm}
\subsection{Simulation Experiments}
\vspace{-0.3cm}
\begin{table}[ht]
\centering
\captionsetup{font=footnotesize}
\caption{Configurations of two simulation environments.}
\label{tab:sim_env}
\renewcommand\arraystretch{1.3} 
\tabcolsep=0.4cm 
\begin{tabular}{c|cl|cl}
\hline
                  & \multicolumn{2}{c||}{\textit{Scenario1}}             & \multicolumn{2}{c}{\textit{Scenario2}}             \\ \hline
Terrain Type      & \multicolumn{2}{c||}{Hilliness}             & \multicolumn{2}{c}{Forest}                \\
Area Size         & \multicolumn{2}{c||}{32.32$\times$32.47m} & \multicolumn{2}{c}{43.76$\times$43.79m} \\
Max Ground Height & \multicolumn{2}{c||}{2.973m}                & \multicolumn{2}{c}{1.426m}                \\
Min Ground Height & \multicolumn{2}{c||}{0.549m}                & \multicolumn{2}{c}{0.582m}                \\ \hline
\end{tabular}
\end{table}
\vspace{-0.2cm}
The developed framework is firstly verified in simulation environments. The simulation experiments are conducted on a laptop with an Intel Core i7-10875H CPU and 16 GB memory. We implement the proposed method upon the ROS Noetic software system \cite{ROS}. The simulator is Gazebo which is a 3D dynamic simulator that can accurately and efficiently simulate robots in complex environmental scenarios. In this simulator, we build a differential-driven robot with a 16-beam LiDAR sensor. 
We establish two simulation environments and their configurations are listed in Tab. \ref{tab:sim_env}.


To verify the effectiveness of the developed framework, we conduct an experiment in the first scenario, as shown in Fig. \ref{fig: whole_process}. The target lies in an unknown area away from the robot and there is no prior map. The trajectory snapshot during the navigation is depicted in Fig. \ref{fig: whole_process}(a)-(e), where the trajectory color indicates the ground height of the trajectory.  As shown in Fig. \ref{fig: whole_process}(f)-(j), the local tree and the global graph can be established successfully. With the developed approach, the robot can reach the target across the unknown and uneven terrain. Besides, it is noteworthy that the trajectory has almost the same color with few drastic color changes, as showcased by Fig. \ref{fig: whole_process}(a)-(e), which illustrates that the generated path is relatively flat. Overall, this experiment demonstrates the effectiveness of the developed approach. 

\begin{figure}[tp]
	\centering
	\setlength{\abovecaptionskip}{-5pt}
	\setlength{\belowcaptionskip}{-10pt}
	\includegraphics[width=0.48\textwidth]{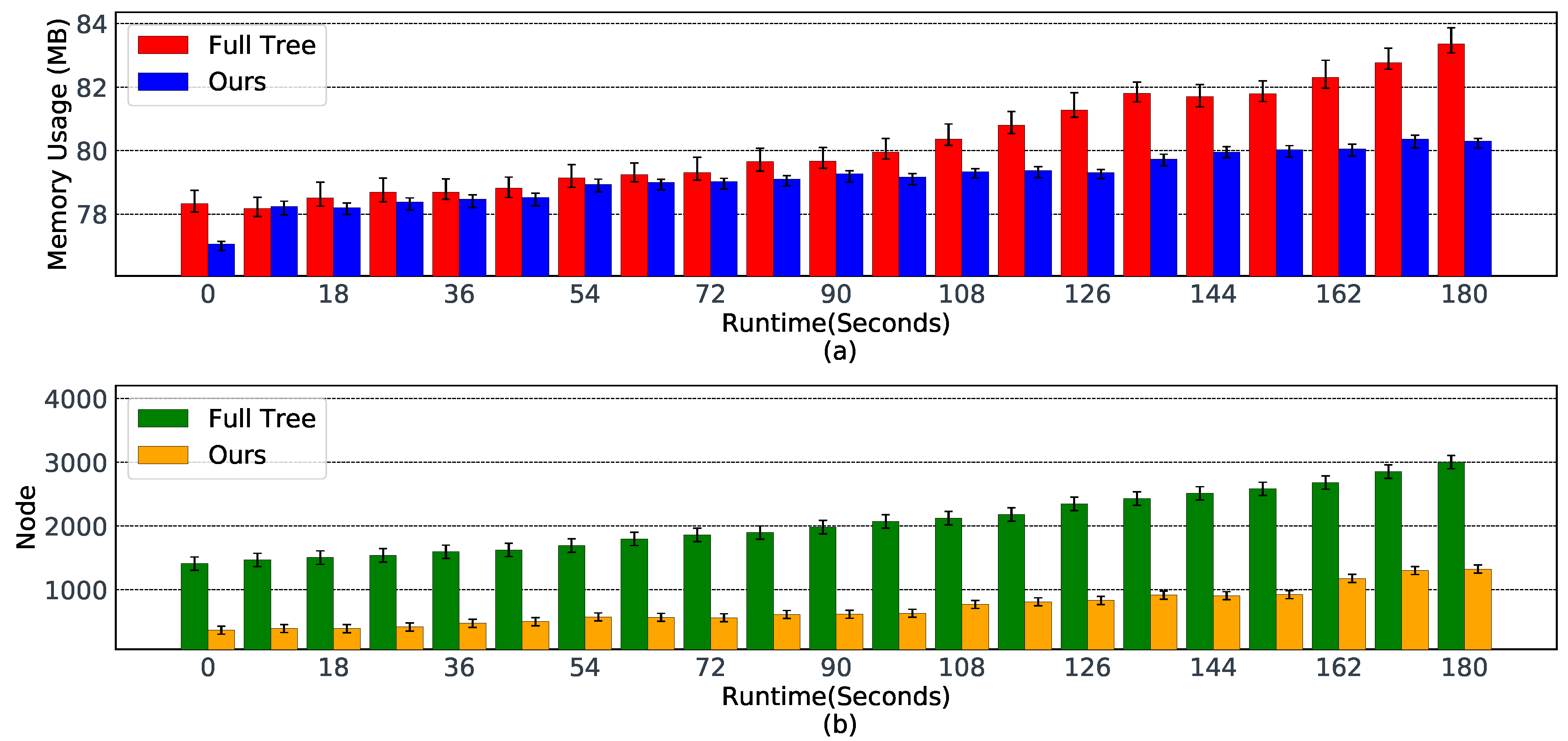}
        \vspace{0.15cm}
	\caption{The comparison of memory consumption and node numbers during the navigation process.}
	\label{fig: compare}
\end{figure}

The layered structure consisting of the graph and tree helps reduce the computational cost. To verify this claim, we compare the developed framework with the \textbf{Full Tree} approach. This approach maintains the whole tree structure, which is a typical routine in the state-of-the-art pipeline \cite{full_tree1}. The other parts of the \textbf{Full Tree} are the same as ours. Five independent experiments are conducted with the same configurations in Fig. \ref{fig: whole_process}.
The results are illustrated in Fig. \ref{fig: compare}. Intuitively, the memory usage of our method maintains the slow growth while that of the \textbf{Full Tree} method exhibits faster rising during the navigation process.  Fig. \ref{fig: compare}(b) shows the comparison of total generated nodes between these two methods. It can be observed that the node growth speed of our method is significantly slower than \textbf{Full Tree} method. 
It demonstrates that the dynamic node pruning mechanism in the tree and graph construction is effective in reducing the total nodes, thus reducing the memory cost in the mapless navigation.

\begin{figure}[tp]
	\centering
	\setlength{\abovecaptionskip}{-5pt}
	\setlength{\belowcaptionskip}{-10pt}
	\includegraphics[width=0.45\textwidth]{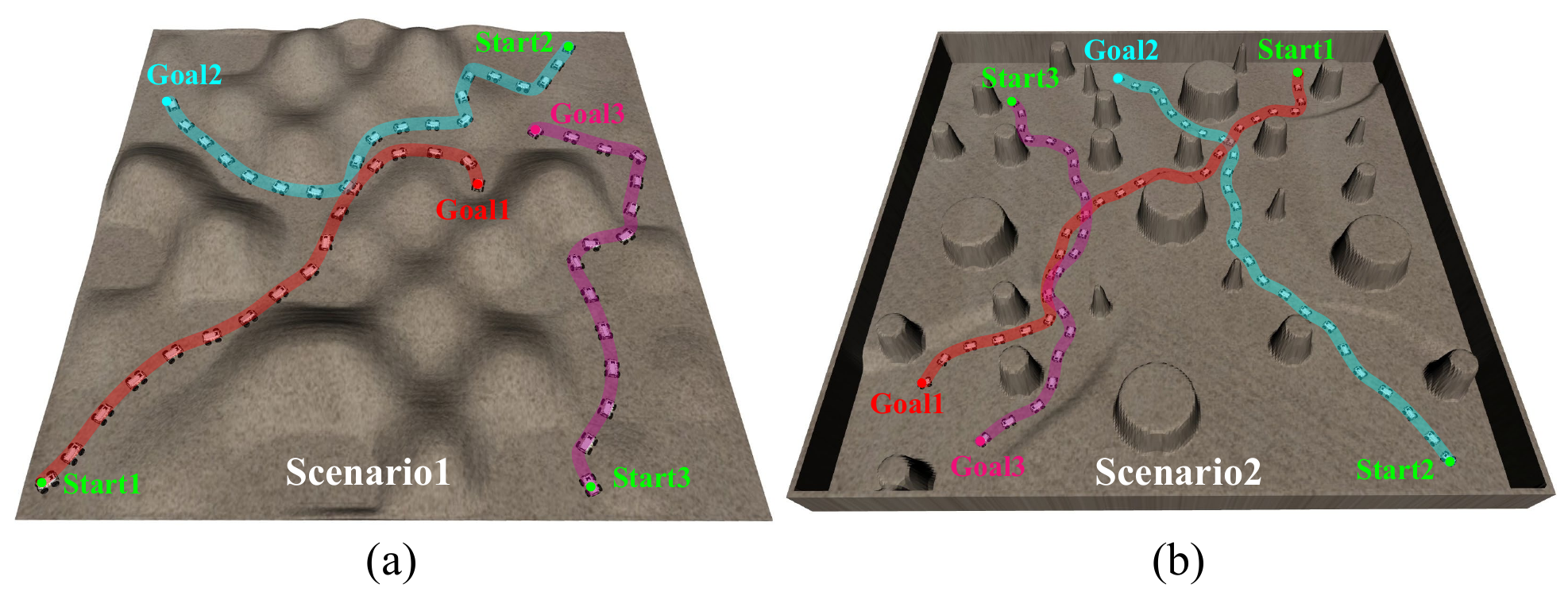}
        \vspace{0.15cm}
	\caption{Evaluations of the developed framework with different configurations in two scenarios.}
        \vspace{-0.2cm}
	\label{fig: robustness}
\end{figure}

To evaluate the robustness of accessibility of the developed approach, we conduct experiments in two different environment scenarios with different start and goal locations. The experimental results are shown in Fig. \ref{fig: robustness}. With any configurations, the robot can successfully find a risk-free route from its current location to the target. 

\vspace{-0.15cm}
\subsection{Experiments in the Real World}
\label{realworld experiment}


To verify the effectiveness of the developed approach, we conduct experiments in real-world environments. 
We use the Agile Scout 2.0 as the robot chassis that is equipped with necessary sensors like Ouster LiDAR OS-32. The computation processor is the NVIDIA Jetson AGX Orin. The differential-driven robot has the capacity to travel at the ground with less than $30^{\circ}$ slope. We use the FAST-LIO2 \cite{fastlio2} for localization.

\begin{figure}[tp]
	\centering
	\setlength{\abovecaptionskip}{-5pt}
	\setlength{\belowcaptionskip}{-10pt}
	\includegraphics[scale=0.375]{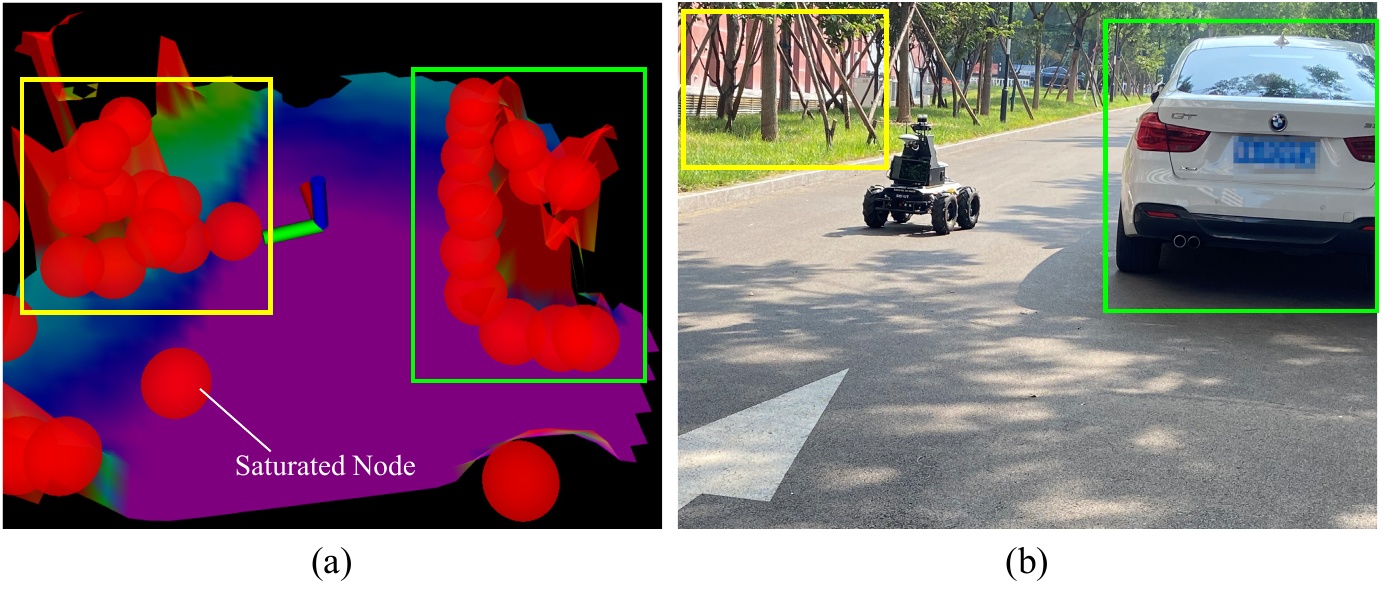}
        \vspace{-0.2cm}
	\caption{Saturated node generation in the real-world environment.}
	\label{fig: real_obs}
\end{figure}

In particular, we look in detail at the performance of saturated node and hazardous region generation mechanisms in the outdoor environment, which is depicted in Fig. \ref{fig: real_obs}. The hazardous regions are clearly marked by the saturated nodes shown as red bubbles in Fig. \ref{fig: real_obs}(a), which illustrates that HD-RRT could avoid expanding inside hazardous regions to guarantee the feasibility of edges. The robot can avoid these regions and achieve risk-free mapless navigation.

\begin{figure}[tp]
	\centering
	\setlength{\abovecaptionskip}{-5pt}
	\setlength{\belowcaptionskip}{-10pt}
	\includegraphics[scale=0.4]{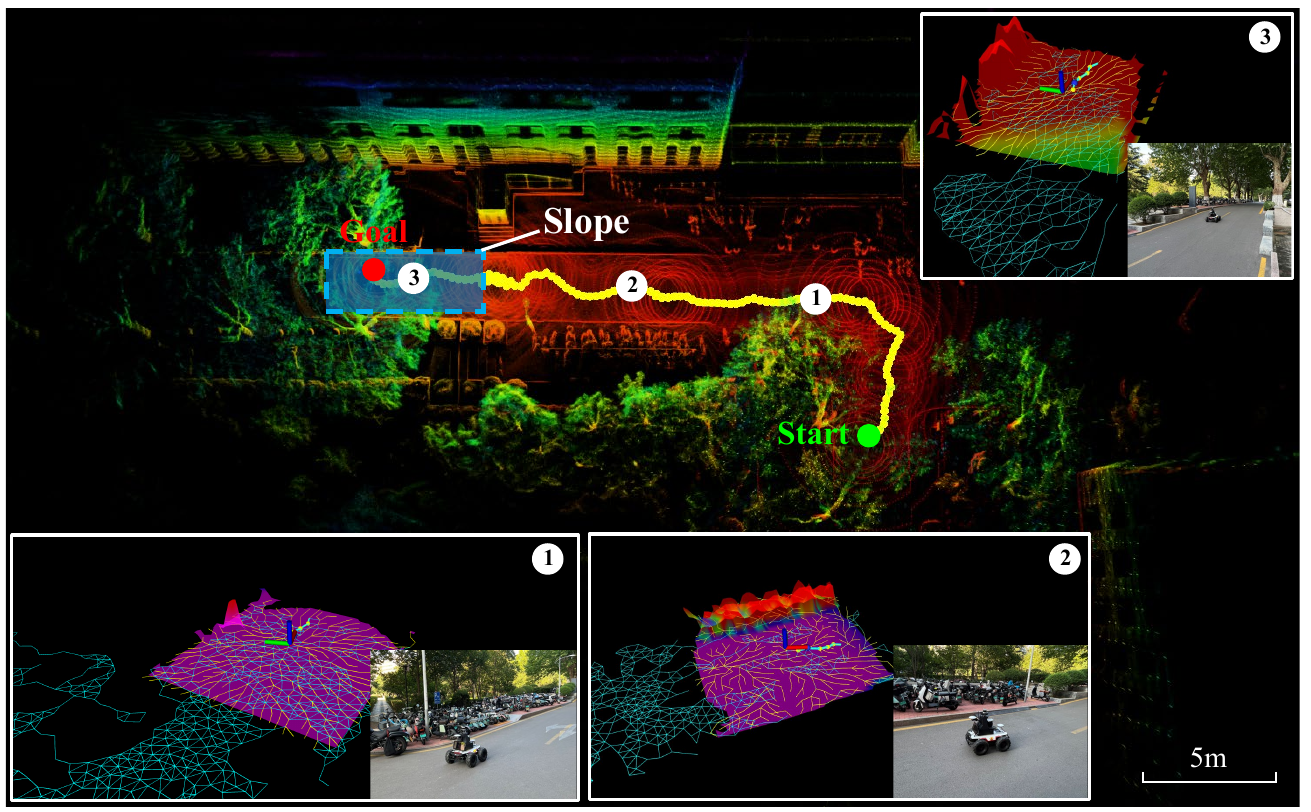}
        \vspace{-0.1cm}
	\caption{The performance of the developed approach in real-world scenario.}
        \vspace{-0.3cm}
	\label{fig: real_uneven}
\end{figure}

The overall performance of the developed approach is shown in Fig. \ref{fig: real_uneven}. The target lies on a slope and there is no prior map. It is observed that with our method, the local tree and the global graph can be generated effectively. The subgoals leading the navigation are evaluated in a timely manner. The robot is able to quickly determine the risk-free regions and finally reach the target, which demonstrates the effectiveness and robustness of the framework in the real-world. 



\section{Conclusions}
\label{conclusion}

In this study, we present a layered structure to achieve mapless navigation in unknown uneven terrain environments. The tree structure can be extended dynamically with the navigation and the hazardous area can be 
marked explicitly. Some nodes of the tree structure are carefully kept and form a graph, which can record the exploration history and contribute to further decision-making process. The subgoals for leading the mapless navigation, together with the route to reach there, can both be queried efficiently on the tree and graph structure. 
Both the simulation and real-world experiments are conducted and the experimental results support the claims of the developed approach. The developed framework is effective in helping the robot reach a target in unknown uneven environments. 
\bibliography{ref.bib}
\balance


\end{document}